\def\year{2020}\relax
\documentclass[letterpaper]{article} 
\usepackage{aaai20}  
\usepackage{times}  
\usepackage{helvet} 
\usepackage{courier}  
\usepackage[hyphens]{url}  
\usepackage{graphicx} 
\usepackage{graphicx}  
\usepackage{framed}
\usepackage{booktabs}
\usepackage{array}
\usepackage{xfrac}
\usepackage{multirow}
\usepackage{colortbl}
\usepackage{bm}
\usepackage{amsfonts}
\usepackage{xspace}
\definecolor{mygray}{gray}{.9}

\makeatletter
\DeclareRobustCommand\onedot{\futurelet\@let@token\@onedot}
\def\@onedot{\ifx\@let@token.\else.\null\fi\xspace}

\def\eg{\emph{e.g}\onedot} 
\def\ie{\emph{i.e}\onedot}

\def\wrt{w.r.t\onedot} 

\makeatother

\title{Softmax Dissection: Towards Understanding Intra- and Inter-class \\ Objective for Embedding Learning}
\author{Lanqing He\thanks{Equal contribution}, Zhongdao Wang$^*$, Yali Li, Shengjin Wang\\ 
Department of Electronic Engineering, Tsinghua University\\
hlq17@mails.tsinghua.edu.cn, wcd17@mails.tsinghua.edu.cn\\
liyali13@mail.tsinghua.edu.cn, wgsgj@tsinghua.edu.cn
}
 
\begin{document}

\maketitle

\begin{abstract}
   The softmax loss and its variants are widely used as objectives for embedding learning applications like face recognition. However,  the intra- and inter-class objectives in Softmax are entangled, therefore a well-optimized inter-class objective leads to relaxation on the intra-class objective, and vice versa. In this paper, we propose to dissect Softmax into independent intra- and inter-class objective (D-Softmax) with a clear understanding. It is straightforward to tune each part to the best state with D-Softmax as objective.Furthermore, we find the computation of the inter-class part is redundant and propose sampling-based variants of D-Softmax to reduce the computation cost.
   The face recognition experiments on regular-scale data show D-Softmax is favorably comparable to existing losses such as SphereFace and ArcFace.
   Experiments on massive-scale data show the fast variants significantly accelerates the training process (such as $64\times$) with only a minor sacrifice in performance, outperforming existing acceleration methods of Softmax in terms of both performance and efficiency.
\end{abstract}

\begin{figure}[t]
\centering
\includegraphics[width=0.9\columnwidth]{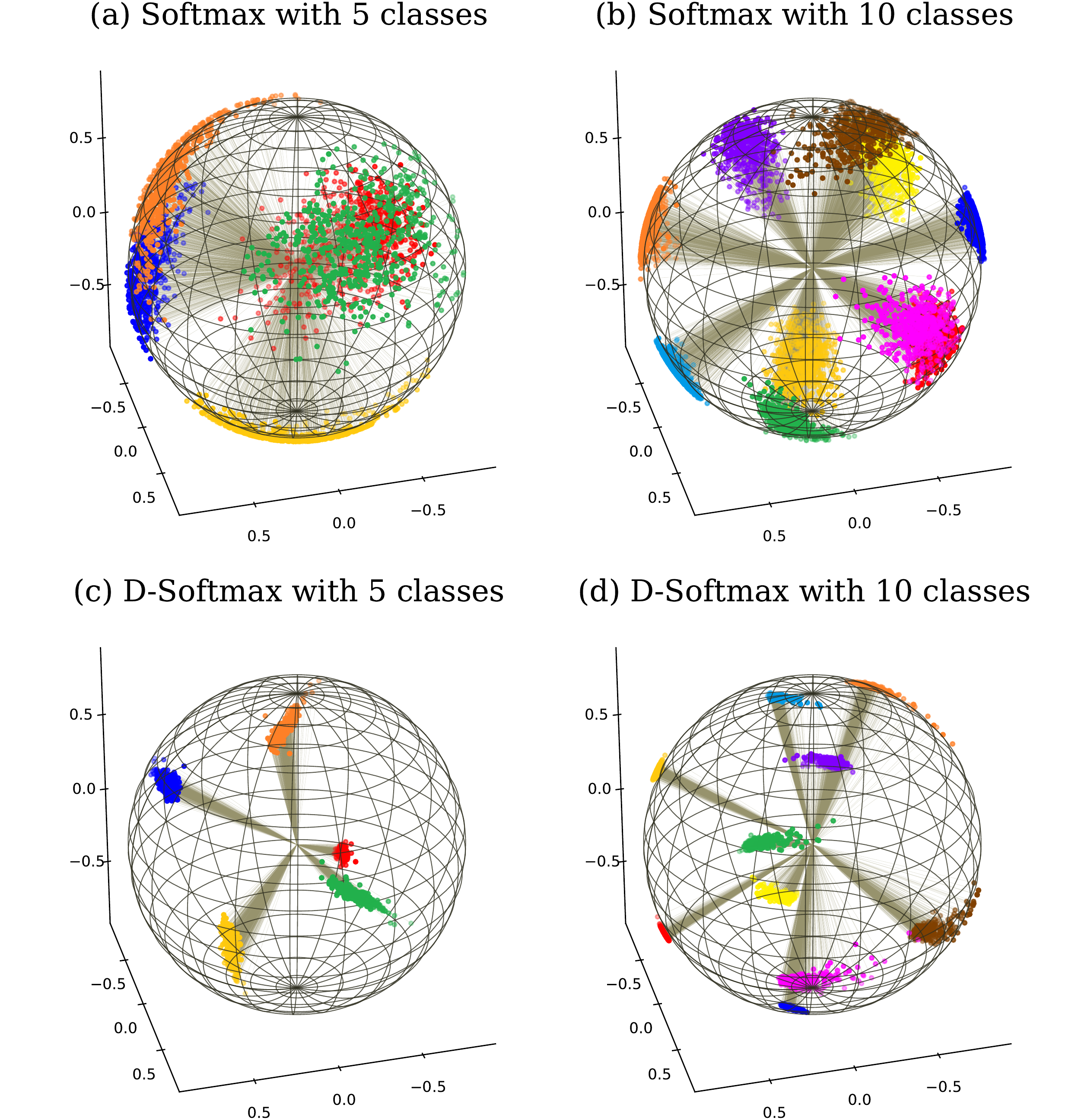}
   \caption{(a),(b): How intra- and inter-class objectives are entangled in Softmax. The inter-class distance in 5-class case is larger than that in 10-class case, therefore the constraint of intra-class objective is relaxed.
    (c),(d): The intra- and inter-class objectives are disentangled in D-Softmax. The intra-class distance is almost the same in both case.}
\label{fig:intro}
\end{figure}

\section{Introduction}
\label{introduction}

Recent years have witnessed the prosperous development of deep learning and its applications.
Among them, embedding learning~\cite{sphereface,cosface,arcface,triplet} (or deep metric learning~\cite{lifted,npair,angular}) is one of the most challenging problems that attracts wide attention, and corresponding research findings are supporting many applications like face recognition and person re-identification~\cite{spherereid,hypershperereid}.

The objective of embedding learning is to learn a mapping function $f(\cdot;\bm{\theta}):\mathcal{X} \rightarrow{\mathbb{R}^{n}}$ so that in the embedding space $\mathbb{R}^{n}$ the distance between similar data is close while the distance between dissimilar data is far. The most straightforward choice is to formulate the embedding learning problem as a classification problem by employing the softmax loss as the objective. 
For instance, in face recognition, faces of different persons are considered as different classes and a large Softmax is used for learning the face embedding.

However, there exist two major drawbacks in the softmax loss. First, the intra- and inter-class objectives are entangled. Such entanglement is visualized in Fig. \ref{fig:intro} (a),(b). We select 5 and 10 identities respectively in the MS-Celeb-1M~\cite{ms1m} dataset with the most samples, set the embedding dimension to $3$ and plot the features. 
One can observe that with large inter-class distance the intra-class distance is also large. As we will show in Sec. \ref{sec:softmaxdissection}, the reason is that the Softmax will gradually relax the intra-class objective along with the increase of inter-class distance, and vice versa.
To our knowledge, we are the first to discuss such entanglement, while existing works mostly address the insufficient discrimination issue by introducing additional  supervision~\cite{triplet,contrastive} or adding angular margin to the softmax loss~\cite{sphereface,cosface,arcface}.

Another shortage should be mentioned is time and memory cost. The softmax loss, as well as its variants, needs to compute class activations over \emph{all} the classes.
This leads to linear time and memory complexity \wrt the number of classes. In practice, the number of classes may be excessively large, say $10^{6}$ or even beyond. 
The excessive memory demand makes it difficult to load all the class weights into the limited GPU memory, and the dramatically increased time cost is also not acceptable. 
Contrastive loss and triplet loss are possible alternatives that do not require much memory, but in terms of accuracy they significantly underperform the softmax family.

In this paper, we propose to dissect the softmax loss into intra- and inter-class objective. The intra-class objective pulls the feature close with the positive class-weight until a pre-defined criterion is satisfied, and the inter-class objective maintains the class weights to be widely separated in the embedding space.
With the dissected softmax (D-Softmax) loss as the optimization objective, the intra- and inter-class objectives are disentangled, so that even the inter-class objective is well-optimized, the constraint on the intra-class objective is still rigorous (Fig. \ref{fig:intro} (c),(d)).

Moreover, D-Softmax also dissects the computation complexity of Softmax into two independent parts. 
The intra-class objective only involves the sample and one positive class-weight, in contrast, the inter-class objective needs to compute activations over \emph{all} negative classes.
We find that such massive computation for the inter-class objective is somehow redundant.
To facilitate the computation, we proposed to sample a \emph{subset} of negative classes in one training pass. 
According to the difference in sampling strategies, we term the lightened D-Softmax as D-Softmax-B and D-Softmax-K respectively. Experiments show both strategies significant accelerate the training process with only a minor sacrifice in performance.

Our major contribution can be summarized as follows:

(1)We propose D-Softmax that dissects the intra- and inter-class objective of the softmax loss. 
The dissected intra-class objective is always rigorous, independent of how well the inter-class objective is optimized, and vice versa. Experiments show D-Softmax is favorably comparable with existing methods such as ArcFace on face recognition task.

(2)We make an important conclusion that the computation of inter-class objective is redundant and propose two sampling-based variants to facilitate the computation of D-Softmax. Training with massive classes (757K),  our methods significantly accelerate the training process with only a minor sacrifice in performance.

\section{Related Work}
\label{relatedwork}
\textbf{Softmax and its variants for face recognition.} It is a widely adopted approach to formulate the face recognition as a multi-class classification problem. DeepFace~\cite{deepface} and DeepID series~\cite{deepid,contrastive,contrastive2} employ the conventional softmax loss in which the class activation is modeled as the inner product between vectors.  
Such loss is not effective enough, and some recent works address this problem by normalizing the embedding~\cite{l2norm} or the class-weights~\cite{weightnorm}. NormFace~\cite{normface} employs to normalize the both, which is equivalent to optimize the cosine distance.
This inspired a series of works on softmax variants that optimize the angular distances between classes by introducing the angular margin~\cite{sphereface,cosface,arcface}.
However, all aforementioned losses focus on strengthening the constraint but overlook a fact, that the insufficiency of Softmax is essentially caused by the entanglement of the intra- and inter-class objective.

\textbf{Acceleration for Softmax.} The acceleration of Softmax is an extensively studied problem typically in natural language processing, where large vocabularies need to be deal with. 
Existing methods mainly re-organize the structure of Softmax by the hierarchy of words~\cite{hierarchy}, or the imbalanced frequency of classes~\cite{langfreq1,langfreq2,langfreq3,langfreq4}. 
However, these methods do not apply to real-world applications like face recognition, because the data are not hierarchical-structured nor substantially imbalanced on importance. HF-Softmax~\cite{hfsoftmax} is a relatively related work to ours. It dynamically selects a subset of the training classes, by constructing a random forest in the embedding space and retrieving the approximate nearest neighbors. The time cost of loss computation is indeed reduced, but the update of the random forest still cost too much time. 
In this work, the light version D-Softmax do not require any extra computation besides the loss itself, so the computation is much faster. Moreover, the dissected intra-class objective is always rigorous, thus the performance is also superior.

\begin{figure*}[t]
\centering
   \includegraphics[width=0.8\textwidth]{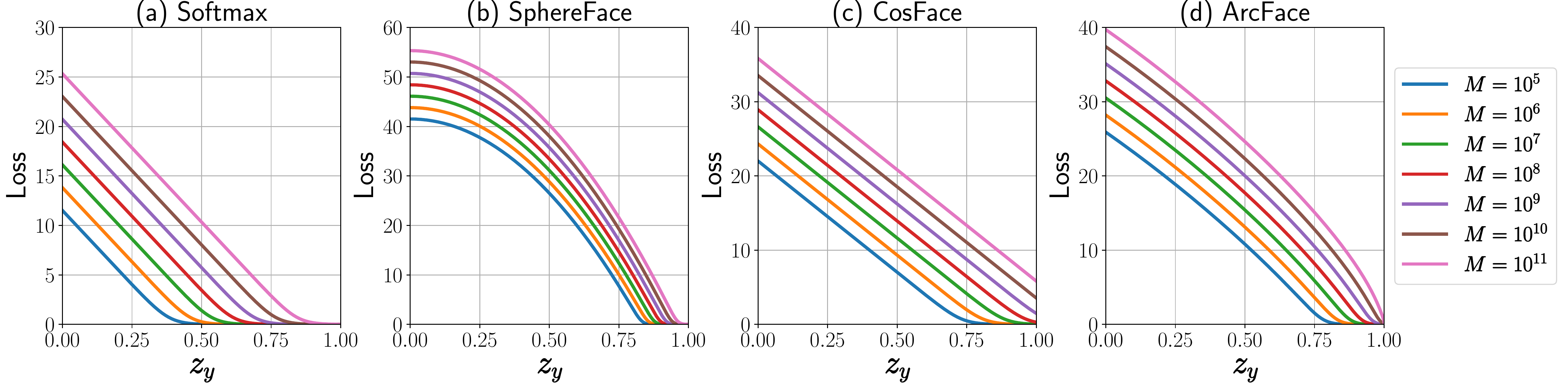}
   \caption{How the loss value varies with fixed inter-class similarity $M$ against varying ground-truth class activation $z_y$ in (a) Softmax, (b) SphereFace, (c) CosFace and (d) ArcFace. Different curves mean different $M$ values.}
\label{fig:intra}
\end{figure*}

\section{Softmax Dissection}
\label{sec:softmaxdissection}

\subsection{Preliminary Knowledge}
The softmax Cross-Entropy loss is fomulated as,
\begin{equation}
    \mathcal{L}_{s} = - \log(\frac{e^{sz_y}}{\sum_{i=1}^K e^{sz_k}}) = \log(1+\frac{\sum_{k\ne y}^K e^{sz_k}}{e^{sz_y}})
\end{equation}
where $s$ is a scale parameter, $z_k$ indicates the activation of the $k$-th class, $k\in\{1,2,...,K\}$ and $K$ is the number of classes. 
We denote the activation of the ground-truth class as $z_y$. 
In conventional Softmax, $z_k = \bm{w}_k^T\bm{x}$,  where $\bm{w}_k$ is the class weight and $\bm{x}$ is the feature of the last fully connected layer. 
In recent arts \eg NormFace~\cite{normface}, the activation is usually modified as $z_k = \cos{(\theta_{\bm{w}_k, \bm{x}})}$. 
We adopt this cosine formulation for its good performance and intuitive geometric interpretation. Here we also list several variants of Softmax, \ie, SphereFace~\cite{sphereface}, ArcFace~\cite{arcface} and CosFace~\cite{cosface},

\begin{equation}
\label{eq:sphere}
    \mathcal{L}_{sphere} = \log(1+\frac{\sum_{k\ne y}^K e^{sz_k}}{e^{s\cos{(m_1 \arccos{z_y})}}})
\end{equation}

\begin{equation}
\label{eq:arc}
    \mathcal{L}_{arc} = \log(1+\frac{\sum_{k\ne y}^K e^{sz_k}}{e^{s[\cos{(\arccos{z_y}) + m_2}]}})
\end{equation}

\begin{equation}
\label{eq:cos}
    \mathcal{L}_{cos} = \log(1+\frac{\sum_{k\ne y}^K e^{sz_k}}{e^{s(z_y-m_3)}})
\end{equation}
where $m_1, m_2$ and $m_3$ are hyperparameters that control the inter-class margin. 
Note that the only difference between these loss functions is the denominator in the fraction.
\subsection{The Intra-Class Component}
\label{sec:intra}
In this section, we first introduce how the intra-class objective is entangled with the inter-class objective. Then we compare the intra-class objective between Softmax and its margin-based variants (Eq.\ref{eq:sphere}-\ref{eq:cos}). Finally, we present the intra-class objective of our Dissected Softmax loss.

Let $M={\sum_{k\ne y}^K e^{sz_k}}$ represent the numerator in the fraction in the loss.
$M$ reflects the inter-class similarity. Large $M$ means that the input has large cosine-similarity with all negative classes. 
With fixed $M$, we plot the loss $\mathcal{L}_{s} =\log(1+\frac{M}{e^{sz_y}})$ against the ground-truth class activation $z_y$ in Fig. \ref{fig:intra} (a). 
Two observations can be made.

First,  this family of curves can be approximated by piecewise linear functions: when $z_y$ is small, $\mathcal{L}_{s} \rightarrow{\log M - sz_y}$, and when $z_y$ is large, $\mathcal{L}_{s} \rightarrow{0}$. It implies that when the intra-class similarity $z_y$ is small, the loss will back-propagate a near-constant gradient, while the gradient is almost $0$ when $z_y$ is large. Second, the inflection point where the gradient changes is positively correlated to $M$.We can figure out the intersection point $d$ of the piecewise linear function:

\begin{equation}
\label{eq:d}
    d = \frac{\log M}{s} 
\end{equation}
$d$ can be considered as an approximate termination point of optimization because the gradient vanishes.

This observation supports an important conclusion:

\begin{itemize}
\item \textbf{Conclusion \#1}:
With Softmax as objective, when the class weights are widely separated (lead to small $M$), the optimization of intra-class objective almost terminates at a small value.
\end{itemize}

Unfortunately, the condition that \emph{the class weights are widely separated}(explained in Sec. \ref{sec:d}) always holds in the training process. 
Therefore, the termination of intra-class objective optimization is always so early that the training is not sufficient. By comparing the loss curves in Fig. \ref{fig:intra}, we speculate the early termination of the intra-class similarity optimization is the main reason why Softmax underperforms its margin-based variants. 
All termination points of variant curves have significant positive shifts compared to the vanilla Softmax under the same $M$.
This means these losses do not stop optimizing the intra-class similarity until $z_y$ is pretty large (say $0.8$).

To address this problem that $M$ is not large enough, we propose to disentangle the intra-class objective from the inter-class objective, by replacing $M$ with a constant value $\epsilon$. 
In this manner, we can manually adjust the optimization termination point $d$ of the intra-class similarity to a sufficiently large value. 
To summarize, the intra-class component of the Dissected Softmax is:
\begin{equation}
\label{eq:intra}
    \mathcal{L}_{D}^{intra} =\log(1+\frac{\epsilon}{e^{sz_y}})
\end{equation}

\subsection{The Inter-Class Component}
\label{sec:inter}
In Sec. \ref{sec:intra} we modified softmax loss and obtain a disentangled intra-class objective. However, we still need inter-class objective as a \emph{regularization} to avoid collapsing to a trivial solution where all the data is mapped to a single point. 
Similarly, we first analyze the inter-class objective of softmax and its variants, then give the formulation of the inter-class objective of D-Softmax.

Consider a sample $x$ of class $y$ and its activation on the $n$-th ($n \ne y$) class $z_n$. Softmax loss can be written as, 
\begin{equation}
    \mathcal{L}_{s} =  \log(1+\frac{e^{sz_n} + \sum_{k\ne y,n}^K e^{sz_k}}{e^{sz_y}}) = \log(1+\frac{e^{sz_n} + M_n}{e^{sz_y}})
\end{equation}
where we  replace the summation with  $M_n$  for convenience. 

\begin{figure}[t]
\centering
   \includegraphics[width=0.9\columnwidth]{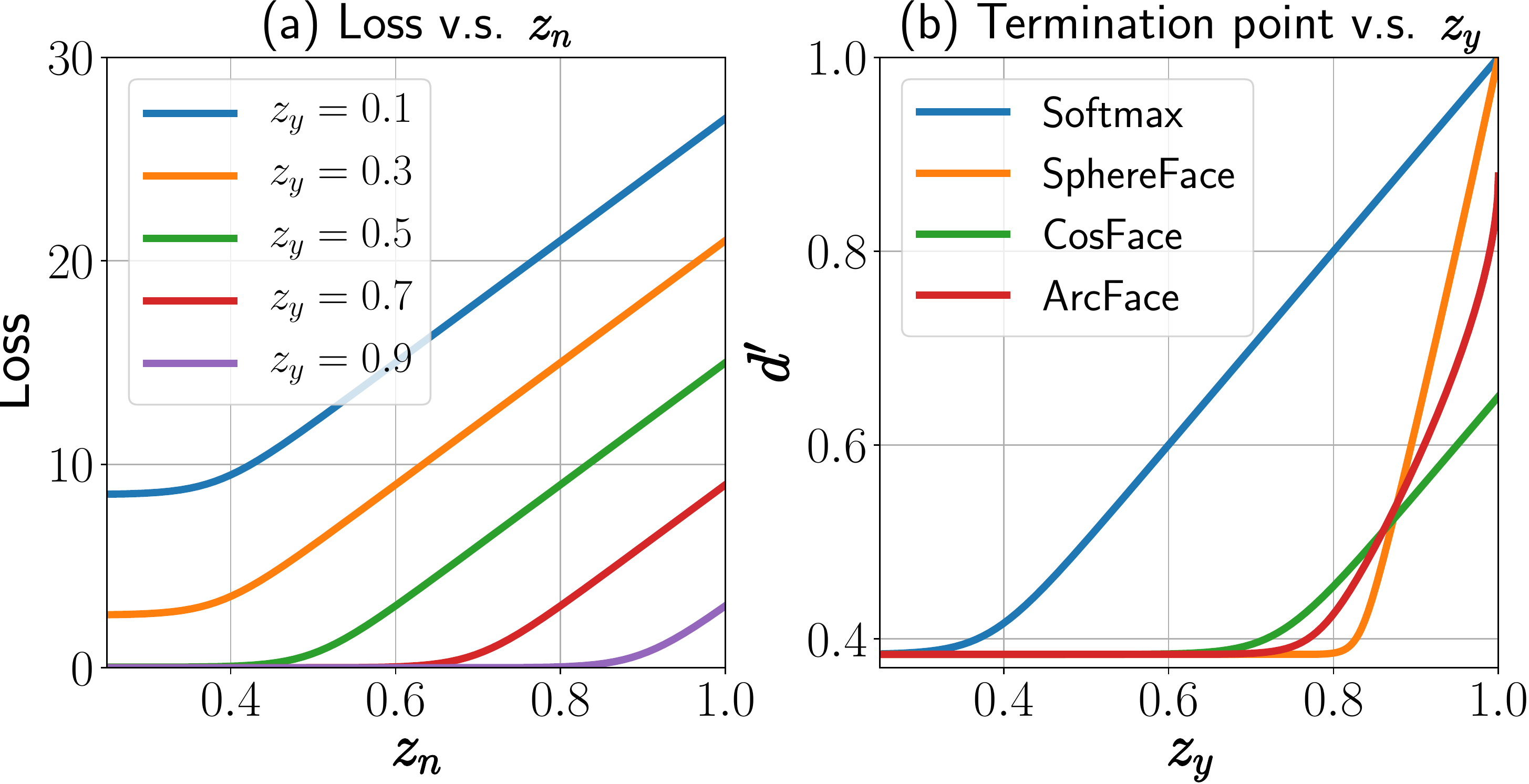}
   \caption{How the inter-class objective of Softmax is entangled with the intra-class objective. 
   (a) The loss curve of Softmax against negative-class similarity $z_n$ under different positive-class similarity $z_y$.
   (b) The termination point of the inter-class objective \wrt varying intra-class similarity $z_y$.}
\label{fig:inter}
\end{figure}

Firstly we fix $M_n$ and study how the loss varies with different $z_n$ and $z_y$. 
A family of curves are presented in Fig. \ref{fig:inter} (a). Similar characteristic emerges like in the intra-class analysis: The gradient $\frac{\partial \mathcal{L}_{s}}{\partial z_n}$ remains almost constant with large negative-class similarity $z_n$ and diminishes rapidly to $0$ at some point. Once again we define the optimization termination point for $z_n$ as the intersection point of the approximate piecewise linear function,

\begin{equation}
    d^{\prime} = \frac{\log(e^{sz_y} + M_n)}{s}    
\end{equation}
and a conclusion can be drawn,

\begin{itemize}
\item  \textbf{Conclusion \#2}:
With Softmax as objective, when the intra-class similarity ($z_y$) is large, the optimization of the negative class weights almost terminates at a large value.
\end{itemize}

This may lead to non-sufficient discrepancy among different class weights thus hamper the embedding learning. 
As an evidence, we plot in Fig. \ref{fig:inter} (b) the termination point $d^{\prime}$ against the intra-class similarity $z_y$ for Softmax, SphereFace, CosFace and ArcFace.
Wider plateau in the curve means the objective regularizes the inter-class similarity more rigorously.
All the large-margin Softmax variants present much wider plateau than the vanilla Softmax. 

In light of above analysis, we propose to disentangle the inter-class objective, by replacing the intra-class similarity $e^{sz_y}$ with a constant. We simply set this constant to 1, therefore the inter-class component of the Dissected Softmax is,
\begin{equation}
\label{eq:inter}
    \mathcal{L}_{D}^{inter} =\log(1+\sum_{k\ne y} e^{sz_k})
\end{equation}
In such manner, the $d^{\prime}$ curve is a flat line, which means the regularization on inter-class similarity is always strict.

\subsection{D-Softmax and Its Light Variants} 
\label{sec:d}
Based on Eq.\ref{eq:intra} and \ref{eq:inter}, the final form of the Dissected Softmax (D-Softmax) loss is,
\begin{equation}
    \mathcal{L}_{D} = \mathcal{L}_{D}^{intra} + \mathcal{L}_{D}^{inter} \\
                 = \log(1+\frac{\epsilon}{e^{sz_y}}) + \log(1+\sum_{k\ne y} e^{sz_k})               
\end{equation}

The merits of Dissected Softmax are mainly two-folds.
First, as we learn from Conclusion $\#1$ and $\#2$ that, in vanilla Softmax, the optimization of intra- and inter-class objective is entangled. Minimizing the intra-class objective will relax the regularization on inter-class objective, and vice versa. In D-Softmax, the optimization is disentangled, thus the constraints are always strict, so the learned embedding is more discriminative. Second, such disentangled formulation allows us to further reduce the computational complexity of the loss function, significantly boosting the training efficiency when the number of classes is tremendous. 

\begin{figure}[t]
\centering
   \includegraphics[width=0.9\columnwidth]{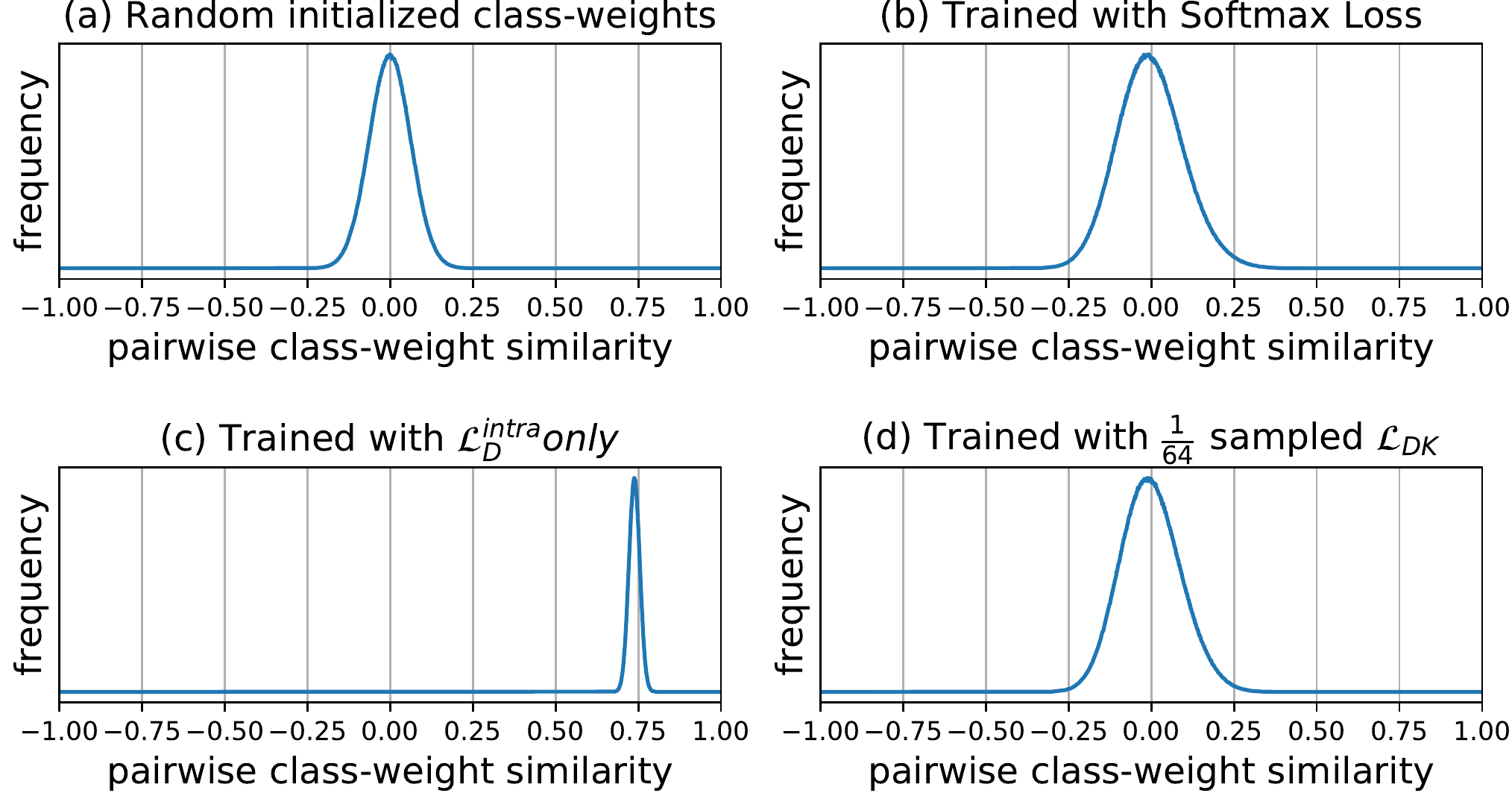}
   \caption{The distributions of pairwise class-weight cosine similarities 
   (a) when the class-weights are randomly initialized, 
   (b) after trained with Softmax,  
   (c) after trained with the intra-class objective $\mathcal{L}_{D}^{intra}$ only,
   (d) after trained with the $\frac{1}{64}$ sampled $\mathcal{L}_{D}^{inter}$.}
\label{fig:distribution}
\end{figure}

\begin{table*}[t]
\small
\centering
\caption{Face recognition performance with different loss functions and performance of D-Softmax with different configurations.
    The best results are \textbf{bolded} and the second best results are \underline{underlined}.}\smallskip
\begin{tabular}{p{2.5cm}<{\centering}| p{0.8cm}<{\centering} |p{1.2cm}<{\centering}p{1.1cm}<{\centering}p{1.2cm}<{\centering}| p{1cm}<{\centering}p{1cm}<{\centering}p{1cm}<{\centering}p{1cm}<{\centering}|p{2.0cm}<{\centering}}
	\toprule
         \multirow{2}{*}{Loss}& \multirow{2}{*}{$d$}  & 
         \multicolumn{3}{c|}{Verification accuracy ($\%$)} & \multicolumn{4}{c|}{IJBC:TAR@FAR ($\%$)}  &
         MegaFace:\\
    \cline{3-9}
    
         & & LFW & CFP & AgeDB & $10^{-1}$  & $10^{-2}$  & $10^{-3}$ & $10^{-4}$ & Rank1@$10^6$ \\
    \hline
    $\mathcal{L}_{s}$
    &-	& 99.30& 87.23	& 94.48& \textbf{98.15}& 95.82 & 91.45 & 85.43 & 91.25 \\
    $\mathcal{L}_{sphere}$
    &- & 99.59 & 91.37 & 96.62 & 98.03& 96.10& 92.60& 86.41 & 96.04\\
    $\mathcal{L}_{arc}$
    &- & 99.68 &92.26& \underline{97.23}&98.01 & 95.84 & 92.64& 87.29 & \underline{96.97} \\
    \hline
    $\mathcal{L}_{D}$ 
    &0.5 & 99.38 &88.34& 95.04&98.02 & 95.80 & 91.45& 85.47 & 92.06\\
    $\mathcal{L}_{D}$ 
    &0.7 & 99.60 &91.44& 96.51&98.08 &\underline{96.15} &92.58 & 86.56 & 95.90\\
    $\mathcal{L}_{D}$ 
    &0.9 & \textbf{99.74} &\underline{92.27}& 97.22&\underline{98.09} & \textbf{96.21} & \underline{92.91} & \underline{88.17} & 96.94 \\
    $\mathcal{L}_{D}$ 
    &1.0 & 99.63 &92.01& 96.88 &98.03 &96.11 & 92.60 & 86.88 & 96.25\\
    \hline
    $\mathcal{L}_{D}^{intra} + \mathcal{L}_{s}^{inter}$ 
    &0.9 & 99.47 & 90.21 & 95.21& 98.01& 95.94 & 91.78& 85.86 & 93.08 \\
    $\mathcal{L}_{D}^{intra} + \mathcal{L}_{arc}^{inter}$ 
    &0.9 & \underline{99.73} & \textbf{93.07} & \textbf{97.30}&98.02 & 96.12 & \textbf{92.97} & \textbf{88.28} & \textbf{97.02} \\
    
	\bottomrule
\end{tabular}
\label{tab:full}
\end{table*}

 When the number of classes is larger than $10^6$, the computation of Softmax becomes the bottleneck of the training process. Denote the batch size as $B$, the number of classes as $K$, then the time complexity for computing Softmax is $\mathcal{O}(BK)$.
In D-Softmax, this complexity is dissected into $\mathcal{O}(B)$ for $\mathcal{L}_{D}^{intra}$ plus $\mathcal{O}(B(K-1))$ for $\mathcal{L}_{D}^{inter}$.
When $K \gg B$, the computation of $\mathcal{L}_{D}^{inter}$ becomes the major time overhead.
In order to accelerate training, let us consider: Is it necessary to compute all the negative-class activations in a mini-batch?

In this work, our answer is \textbf{No}. The main reason lies in the approximate orthogonality of class-weights in high-dimensional space.
For illustrating the approximate orthogonality, we randomly initialize $10,000$ class weights with $256$ dimension and plot how the pairwise cosine similarities distribute in Fig. \ref{fig:distribution}
 (a). The pairwise cosine similarities present a narrow Gaussian distribution with zero mean and around $3\sigma=0.2$, which means the class weights are far apart from each other. We also plot how this distribution changes after training with softmax in Fig. \ref{fig:distribution} (b). 
 Interestingly, the mean of the Gaussian distribution does not shift, and the  variance just increases a little. This means  $\mathcal{L}_{D}^{inter}$ is not pushing the class weights strictly further from each other.
 Considering above two points, we may reach the following conclusion,
\begin{itemize}
\item  \textbf{Conclusion \#3}:
When optimizing in high-dimensional embedding space, the function of the inter-class objective is not pushing class-weights strictly further apart, but mainly maintaining the approximate orthogonality of the class-weights as a regularization.
\end{itemize}

Based on this conclusion, we speculate the $\mathcal{O}(B(K-1))$ computation of $\mathcal{L}_{D}^{inter}$ may be redundant. 
To validate it, we again train an identical model using $\mathcal{L}_{D}^{intra}$ and a sampled $\mathcal{L}_{D}^{inter}$. 
In each mini-batch we randomly sample $\frac{1}{64}$ of the $K-1$ classes as the negative classes. 
After training , we plot the distribution of pairwise cosine similarities between class weights in Fig. \ref{fig:distribution} (d). 
As expected, the distribution is almost the same as training with the full Softmax. 
In Sec. \ref{sec:exp} we will present the performance degradation of the sampled loss compared to the full D-Softmax is also minor while the computation of $\mathcal{L}_{D}^{inter}$ is $64\times$ faster. 
We name this light variant of D-Softmax as D-Softmax-K for the negative classes are sampled from the $K-1$ classes. 
Formally, The mini-batch version of D-Softmax-K is
\begin{equation}
    \mathcal{L}_{DK}  =\sum_{i=1}^{B} \log(1+\frac{\epsilon}{e^{sz_{y_{i}}}}) + \sum_{i=1}^{B} \log(1+\sum_{k\in \mathcal{S}_{K}} e^{sz_k})             
\end{equation}
where $\mathcal{S}_{K} = \{k|k=1,2,...,K\} \backslash \{y_{i}|i=1,2,...,B\}$ means a subset of the class-weight set . 
The sampling rate remains a hyperparameter for performance-speed trade-off. 

An alternative sampling strategy is sampling from mini-batch samples, and we name such strategy as D-Softmax-B,
\begin{equation}
    \mathcal{L}_{DB}  =\sum_{i=1}^{B}  \log(1+\frac{\epsilon}{e^{sz_{y_{i}}}}) + \sum_{i\in \mathcal{S}_{B}}\log(1+\sum_{k=1}^{K} e^{sz_k})             
\end{equation}
where $\mathcal{S}_{B}$ is a subset of batch samples.
The strengths and weaknesses of each strategy will be shown in Sec.~ \ref{sec:exp:light}.

\begin{table*}[t]
\small
    \centering
    \caption{Comparison of D-Softmax-B, D-Softmax-K and other sampling-based Softmax variants in terms of face recognition accuracy. The best results at $\sfrac{1}{64}$ sampling rate are \textbf{bolded}, and the second best results are \underline{underlined}.}
    \begin{tabular}{p{1.9cm}| p{0.8cm}<{\centering} |p{0.8cm}<{\centering} |p{1.1cm}<{\centering} |p{1.0cm}<{\centering}p{0.9cm}<{\centering}p{1.0cm}<{\centering}| p{0.9cm}<{\centering}p{0.9cm}<{\centering}p{0.9cm}<{\centering}p{0.9cm}<{\centering}|p{1.5cm}<{\centering}}
    \toprule
         \multirow{2}{*}{Loss}& \multirow{2}{*}{$\mid\mathcal{S}_{B}\mid$}  & \multirow{2}{*}{$\mid\mathcal{S}_{k}\mid$}  & {Sampling}  & \multicolumn{3}{c|}{Verification accuracy ($\%$)} & \multicolumn{4}{c|}{IJB:TAR@FAR ($\%$)} &
         MegaFace:\\
    \cline{5-11}
         & & &Rate &\footnotesize{LFW} & \footnotesize{CFP} & \footnotesize{AgeDB} & $10^{-1}$  & $10^{-2}$  & $10^{-3}$ & $10^{-4}$ & Rank1@$10^6$ \\

    \hline
    D-Softmax-B & 256 & 85K & 1 & 
    99.74 &92.27& 97.22&98.09 & 96.21 & 92.91 & 88.17 & 96.94\\
    D-Softmax-B & 64 & 85K & \sfrac{1}{4} & 
    99.75 &92.27& 97.18&98.08 & 96.22 & 92.90 & 88.13 & 96.93\\
    D-Softmax-B & 16 & 85K & \sfrac{1}{16} & 
    99.74 &92.24& 96.92&98.03 & 96.20 & 93.02 & 87.98 & 96.91\\
    \rowcolor{mygray}
    D-Softmax-B & 4 & 85K & \sfrac{1}{64} & 
    \textbf{99.60} &\textbf{90.89}&\textbf{ 95.84}&\underline{98.09} & \textbf{95.87} & \textbf{92.15} & \textbf{86.74} & \textbf{95.34}\\
    D-Softmax-B & 1 & 85K & \sfrac{1}{256} & 
    99.50 &89.09& 94.57&97.95 & 95.29 & 91.25 & 85.24 & 91.53\\ 
    \hline
    \rowcolor{mygray}
    D-Softmax-K & 256 & 1.3K & \sfrac{1}{64} & 
    \underline{99.55} &\underline{89.77}& \underline{95.02}&\underline{98.09} & \underline{95.40} & \underline{92.01} & \underline{86.03} & \underline{94.72}\\ 
    Rand-Softmax & 256 & 1.3K & \sfrac{1}{64} & 
    99.07 &85.47& 89.35&98.05 & 94.30 & 87.52 & 78.96 & 88.27\\
    Rand-ArcFace & 256 & 1.3K & \sfrac{1}{64} & 
    99.43 &88.21& 84.08&\textbf{98.11} & 95.14 & 91.27 &84.26&93.62\\
    HF-Softmax & 256 & 1.3K & \sfrac{1}{64} & 
    99.18 &86.11& 91.55&97.92 & 94.45 & 89.63 &81.85&91.18\\
    \bottomrule
    
    \end{tabular}
    \label{tab:light}
     \vspace{-0.2cm}
\end{table*}

\section{Experimental Results}
\label{sec:exp}
\subsection{Datasets and Evaluation Metrics}
\noindent\textbf{Evaluation.}We validate the effectiveness of the proposed D-Softmax in the face recognition task. 
The testing datasets include LFW~\cite{lfw}, CFP-FP~\cite{cfp}, AgeDB-30~\cite{agedb}, IJB-C~\cite{ijbc} and MegaFace~\cite{megaface}.
LFW is a standard face verification benchmark that includes 6,000 pairs of faces, and the evaluation metric is the verification accuracy via 10-fold cross validation.
CFP-FP and AgeDB-30 are similar to LFW but emphasis on frontal-profile  and cross-age face verification respectively. 
IJB-C is a large-scale benchmark for template-based face recognition. 
A face template is composed of multiple face images or video face tracks. Features are simply average pooled in a template to obtain the template feature. 
The evaluation metric is the true accept rate (TAR) at different false alarm rate (FAR). MegaFace identiﬁcation challenge is a large-scale benchmark to evaluate the performance at the million distractors. We perform the rank-1 identification accuracy with $10^6$ distractors on the a refined version used by ArcFace\footnote{https://github.com/deepinsight/insightface}.

\noindent\textbf{Training.} We adopt the MS-Celeb-1M~\cite{ms1m} dataset for training. 
Since the original MS-Celeb-1M contains wrong annotations, we adopt a cleaned version that is also used in ArcFace. The cleaned MS-Celeb-1M consists of around 5.8M images of 85K identities. 
Moreover, to validate the effectiveness and efficiency of the proposed losses on massive-scale data, we combine  MS-Celeb-1M with the MegaFace2~\cite{megaface2} dataset to obtain a large training set. 
The MegaFace2 dataset consists of  4.7M images of 672K identities, so the joint dataset has 9.5M images of 757K identities in total.

\subsection{Experiments on D-Softmax}
\label{sec:exp:full}
In this section, we explore how to set the intra-class termination point $d$ for best performance, and how different formulations of inter-class objective affect the discrimination of the learned embedding. Finally we compare D-Softmax with other state-of-the-art loss functions.

All the models are standard ResNet-50~\cite{resnet}, trained on MS-Celeb-1M. We set the scale $s=32$, the margin $m_1=4$ for SphereFace, $m_2=0.5$ for ArcFace for the best performance. The other hyperparameters are the same.

\begin{figure}[t]
\centering
   \includegraphics[width=0.8\columnwidth]{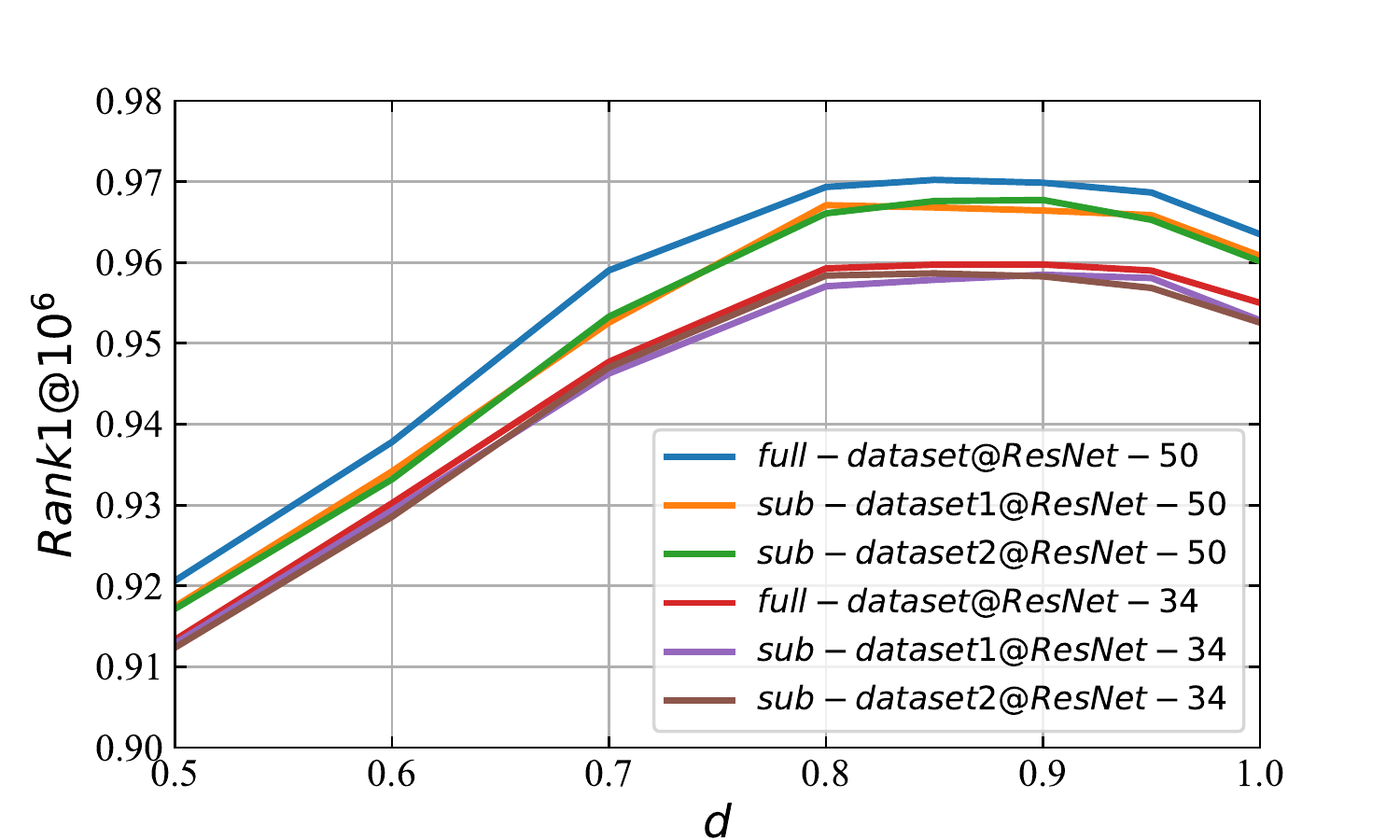}
   \caption{Rank-1 identification accuracy against $10^6$ distractors on MegaFace dataset (refined) with different hyperparameter.We randomly split the full MS-Celeb-1M dataset into sub-dataset1 and sub-dataset2}
\label{fig:rank1}
\end{figure}

\noindent \textbf{Selection of $d$}. By tuning the hyperparameter $\epsilon$ in $\mathcal{L}_{D}$, we are able to set the optimal $d$. Table \ref{tab:full} shows performance of $\mathcal{L}_{D}$ with different settings of $d$. 
With $d$ increasing from $0.5$ to $0.9$, the performance increases steadily. 
However, when we further insrease $d$ to $1.0$, the performance drops slightly. We also perform a range of experiences with different backbone networks and training set and Fig.\ref{fig:rank1} shows how the identiﬁcation accuracy varies with different d. In each setting, the conclusion is consistent: A moderately large termination point for intra-class similarity \eg $0.9$, leads to the best results, so we set $d=0.9$ in all the following experiments.

\noindent \textbf{Different forms of inter-class objective.} Apart from the simple form $\mathcal{L}_{D}^{inter}$ proposed in Sec. \ref{sec:inter}, we also compare  several different forms of inter-class objective, the inter-class objective of Softmax(NormFace) and ArcFace. We donate such objectives as $\mathcal{L}_{s}^{inter}$ and $\mathcal{L}_{arc}^{inter}$.

To accomplish such objectives, in the forward pass we compute the full Softmax or ArcFace loss, while in the backward pass we only back-propagate the inter-class part of the gradients, by setting $\frac{\partial \mathcal{L}}{\partial z_{y}}$ to $0$.
Then we combine $\mathcal{L}_{s}^{inter}$ or $\mathcal{L}_{arc}^{inter}$ with the intra-class part of D-Softmax $\mathcal{L}_{D}^{intra}$ to train a model.
Table \ref{tab:full} compares the performance between $\mathcal{L}_{D}$, $\mathcal{L}_{D}^{intra} + \mathcal{L}_{s}^{inter}$ and $\mathcal{L}_{D}^{intra} + \mathcal{L}_{arc}^{inter}$.
With the same intra-class objective, it is shown that $\mathcal{L}_{D}^{inter}$ outperforms $\mathcal{L}_{s}^{inter}$ by a large margin.
$\mathcal{L}_{D}^{inter}$ and $\mathcal{L}_{arc}^{inter}$ lead to almost the same good performance, which is as expected. Though entangled with the intra-class objective, the regularization on inter-class similarity in ArcFace is rigorous enough until the intra-class similarity is pretty large (say $>0.8$).
Similarly, the proposed dissected form of inter-class objective is always rigorous regardless of the intra-class similarity. Compared with $\mathcal{L}_{arc}^{inter}$, $\mathcal{L}_{D}^{inter}$ has a more concise form with no hyperparameter and is easier to be extended to fast variants.

\begin{table*}[t]
\small
    \centering
    \caption{Comparison between D-Softmax-K and several baseline methods on large-scale training set. The loss/total average time is computed as the average time for one forward-backward pass of the loss layer / the entire model. The best results are \textbf{bolded} and the second best results are \underline{underlined}.}
    \begin{tabular}{p{2.0cm}| c |c  |p{1.0cm}<{\centering}p{0.9cm}<{\centering}p{1.0cm}<{\centering}| p{0.9cm}<{\centering}p{0.9cm}<{\centering}p{0.9cm}<{\centering}p{0.9cm}<{\centering}|p{1.5cm}<{\centering}}
    \toprule
         \multirow{2}{*}{Loss   }& {Loss Avg. }  & {Total Avg.}   & \multicolumn{3}{c|}{Verification accuracy ($\%$)} & \multicolumn{4}{c|}{IJBC:TAR@FAR ($\%$)} &
         MegaFace:\\
    \cline{4-10}
         & {Time (s)}& { Time (s)}&\footnotesize{LFW} & \footnotesize{CFP} & \footnotesize{AgeDB} & $10^{-1}$  & $10^{-2}$  & $10^{-3}$ & $10^{-4}$ & Rank1@$10^6$ \\
         
    \hline

    \rowcolor{mygray}
    Softmax & 3.12 & 3.96 & 
    \underline{99.38} &\underline{87.96}& \textbf{95.60} &\textbf{98.14} & \textbf{95.84} & \underline{91.55} &\underline{85.79}&\underline{93.03}\\
    Rand-Softmax & \textbf{0.20} & \textbf{1.04} & 
    99.10 &85.56& 89.58&98.06 & 94.44& 88.02 & 79.21 & 88.92\\
    \rowcolor{mygray}
    HF-Softmax & 2.04 & 2.88 & 
    99.27 &86.10& 91.82&98.04 & 94.71 & 90.26 &82.23 &91.79\\
    D-Softmax-K & \underline{0.21} & \underline{1.05} & 
    \textbf{99.47} &\textbf{89.59}& \underline{95.32}& \underline{98.10}& \underline{95.66} & \textbf{91.83} & \textbf{85.83} & \textbf{94.54}\\
    \bottomrule
    
    \end{tabular}
    \label{tab:large}
\end{table*}

\noindent \textbf{Comparison with state-of-the-art losses.} For fair comparison, we re-implement NormFace~\cite{normface}, SphereFace and ArcFace and compare the proposed D-Softmax with them using the same training data and model. 
As shown in Table \ref{tab:full}, the proposed D-Softmax outperforms the Softmax (NormFace) baseline even with a small $d=0.5$, and with $d=0.9$ D-Softmax outperforms Softmax by a siginificant margin.
SphereFace and ArcFace also outperform the Softmax baseline because of the introduced angular margin. To tell the difference between $d$ and margin parameteter, we take ArcFace for example. The hyperparameter $m$ is the required angular margin between sample features and negative class weights, affects both the intra- and inter-class objective, so that for all class the intra-class constraint is not the same rigorous. Therefore, it needs to be tuned. The best selection may vary with other hyperparameters varies.
 Instead, we introduce $d$ which has a more clear interpretation to reach the same goal of adding margin. $d$ indicates the optimization termination of distance between sample features and positive class weights. Therefore, it is straightforward to select a reasonable value of d, and for all class the intra-class constraint is the same rigorous. The best selection of d in D-Softmax consistently ranges from 0.8 to 0.9 with different training settings, even when we down-sample the inter-class objective in the next section.

\subsection{Experiments on Light D-Softmax}
\label{sec:exp:light}

In Sec. \ref{sec:d} we proposed two sampling-based variants of D-Softmax, \ie, D-Softmax-B and D-Softmax-K, for reducing the computational complexity of training with massive classes. 
In this section, we explore the strength and weakness of each sampling strategy.

\noindent \textbf{D-Softmax-B}. D-Softmax-B is a most easy-to-implement sampling method for reducing the complexity of the inter-class objective.
In practice, one only needs to sample from the batch samples and then computes all the negative-class activations \wrt the sampled batch samples. 
To illustrate the effectiveness of D-Softmax-B, we train several ResNet-50 with batch size of $256$, and employ D-Softmax-B as the objective, with sampling rates varying from $1$ to $\sfrac{1}{256}$. 
As shown in Table~\ref{tab:light}, the performance drops slowly until the sampling rate is lower than $\sfrac{1}{64}$. The accuracy drop of $\sfrac{1}{16}$ sampling rate is nearly neglectable compared to the non-sampled version, while the computation of $\mathcal{L}_{D}^{inter}$ is $16\times$ faster.
However, even with the extreme sampling rate $\sfrac{1}{256}$, \ie, only one batch sample is used for computing the inter-class objective, the performance of D-Softmax-B is still acceptable($99.50\%$v.s. full-computed version $99.74\%$ LFW accuracy). 
These results in turn strongly support Conclusion \#3 we made in Sec. \ref{sec:d}, that the inter-class objective is mainly maintaining the approximate orthogonality of class weights as a regularization, thus the full-computation with $\mathcal{O}(B(K-1))$ is redundant. 
The advantages of D-Softmax-B are the simplicity for implementation and minimal sacrifice of performance. 
However, it faces a dilemma in practice, \ie, the memory limit of GPU is also a matter in large-scale training.
The computation of D-Softmax-B requires the whole class-weight matrix to be copied to the GPU memory thus adds difficulties on parallelism.

\noindent \textbf{D-Softmax-K}. For each mini-batch, D-Softmax-K first samples candidate negative classes from the intersection of negative-classes sets \wrt every batch sample, then the batch inter-class objective is computed with simple data parallel. 
To tackle the problem of GPU memory limit, inspired by~\cite{hfsoftmax}, we adopt a parameter server to store all the class weights on a large-capacity memory (\textit{e.g.} CPU Ram). When some classes are sampled in a mini-batch, the weights of these classes are retrieved on the parameter server and then cached in the client's GPU. In such manner the dilemma of GPU memory limit is mitigated, and also the implementation is not so complicated. 

However, compared with D-Softmax-B at the same sampling rate (see the gray rows in Table \ref{tab:light}), performance of D-Softmax-K is slightly inferior. A possible interpretation is that in D-Softmax-B all the class weights are updated in every mini-batch thus the class weights are more up-to-date in each iteration.
This suggests sampling from the batch samples can achieve better performance. 
Nevertheless, considering the difference in performance is minor while D-Softmax-K is much easier for parallelism, we suggest to use D-Softmax-K in large-scale training.

\noindent \textbf{Compared with other sampling-based methods.} In order to demonstrate the benefits of D-Softmax, we also compare with some exsiting sampling-based methods. 
The first is random Softmax, which means for one mini-batch the to-be-computed class weights are randomly sampled.
The second is random ArcFace, which is similar to Rand-Softmax but the loss function is ArcFace. 
At the same $\sfrac{1}{64}$ sampling rate, both D-Softmax-B and D-Softmax-K outperform Rand-Softmax and Rand-ArcFace by a significant margin ($99.60/99.55\%$ v.s. $99.07/99.43\%$ LFW accuracy). The sampling operation reduce the inter-class objective, which relaxes the intra-class constraint of Softmax and ArcFace. As for D-Softmax, the intra-class constraint is not affected.

HF-Softmax proposed in ~\cite{hfsoftmax} also needs to be comapred, so we adopt the code released by the authors and train HF-Softmax on the same  dataset for fair comparison.
It also samples from $K-1$ negative classes to reduce the computational cost. 
The difference is that the sampling is not random, they build a hash forest to partition the weight space and find approximate-nearest-neighbor (ANN) class weights for batch samples. As shown in Table \ref{tab:light}, HF-Softmax outperforms Rand-Softmax ($99.18\%$ v.s. $99.07\%$ LFW accuracy), since the negative class weights are sampled from the 'hard-negatives' which are more valuable for optimization.
But compared with D-Softmax, the performance is inferior. It is again because the entanglement between the intra- and inter-class objective. Though hard negative class weights are mined, only the inter-class regularization is improved. The intra-class constraint is still not strict enough.

\noindent \textbf{Large-scale experiments.} In order to validate the acceleration effectiveness of the proposed D-Softmax-K on training, we perform a large-scale experiment on the joint dataset of MS-Celeb-1M and MegaFace2. 
Performance and average time cost of some baseline methods are listed in Table \ref{tab:large}. 
The sampling rate is set to $\sfrac{1}{64}$ in all losses.
HF-Softmax and D-Softmax outperform Rand-Softmax at the same sampling rate in terms of accuracy, yet only D-Softmax outperforms the full Softmax loss. 
Sampling based on the entangled form of Softmax, the performance upper bound of HF-Softmax is comaprable to the full Softmax.
In contrast, the sampled D-Softmax has the ability to exceed full Softmax. 

In terms of the time cost, it is obvious that the full Softmax is the slowest one, with $3.12$s average time cost on the loss layer for one forward-backward pass, while Rand-Softmax is the fastest with $0.20$s.
HF-Softmax is supposed to be efficient because only a small fraction of the weights need to be computed, but the update of the random forest cost too much time ($1.83$s on average, while the computation of loss is only $0.21$s.). This time cost can be decreased by changing to fast ANN algorithm or enlarging the updating time duration of the random forest, but the performance will decrease meanwhile. 
In contrast, the proposed D-Softmax-K provides a pretty good performance-speed trade-off. 
The training with D-Softmax-K is as fast as Rand-Softmax since we do not need to build and update a random forest.

Note that the results of large-scale experiments seem to be inferior to that of training with MS-Celeb-1M alone. 
This is because the MegaFace2 dataset is rather noisy. If trained with a cleaned large-scale dataset,  the performance is supposed to be better.

\section{Conclusion}
In this paper, we propose to dissect the softmax loss into independent intra- and inter-class objectives. By doing so, the optimization of the two objectives is no longer entangled with each other, and as a consequence it is more straightforward to tune the objectives to be consistently rigorous during the training time. The propsed D-Softmax shows good performance in the face recognition task. By sampling the inter-class similarity, it is easy to be extended to fast variants (D-Softmax-B and D-Softmax-K) that can handle massive-scale training. 
We show that the fast variants of D-Softmax significantly accelerate the training process, while the performance drop is quite small.

{\small
\bibliographystyle{aaai}
\bibliography{AAAI-HeLQ.5003}
}

\end{document}